%% file: main.tex
\documentclass[10pt,twocolumn,letterpaper]{article}
\usepackage{graphicx}
\graphicspath{ {./images/} }
\usepackage{cvpr}              


\definecolor{cvprblue}{rgb}{0.21,0.49,0.74}
\usepackage[pagebackref,breaklinks,colorlinks,allcolors=cvprblue]{hyperref}

\def\paperID{45} 
\def\confName{CVPR}
\def\confYear{2025}

\title{From Broadcast to Minimap: Achieving State-of-the-Art SoccerNet Game State Reconstruction}

\author{
    Vladimir Golovkin\thanks{Correspondence to: \texttt{vladimir.golovkin@constructor.tech}} \quad Nikolay Nemtsev \quad Vasyl Shandyba \quad Oleg Udin \\
    Nikita Kasatkin \quad Pavel Kononov \quad Anton Afanasiev \quad Sergey Ulasen \quad Andrei Boiarov \\[0.5em]
    Constructor Tech, Sofia \\
}

\begin{document}
\maketitle
\input{sec/0_abstract}    
\input{sec/1_intro}

\input{sec/2_related_works}
\input{sec/3_methodology}

\input{sec/4_datasets}
\input{sec/5_experiment_setting}
\input{sec/6_Evaluation}
\input{sec/7_Conclusion}
\input{sec/8_Acknowledgments}
{
    \small
    \bibliographystyle{ieeenat_fullname}
    \bibliography{main}
}

\input{sec/X_suppl}


\end{document}

%% file: sec/0_abstract.tex
\begin{abstract}
Game State Reconstruction (GSR), a critical task in Sports Video Understanding, involves precise tracking and localization of all individuals on the football field—players, goalkeepers, referees, and others—in real-world coordinates. This capability enables coaches and analysts to derive actionable insights into player movements, team formations, and game dynamics, ultimately optimizing training strategies and enhancing competitive advantage. Achieving accurate GSR using a single-camera setup is highly challenging due to frequent camera movements, occlusions, and dynamic scene content. In this work, we present a robust end-to-end pipeline for tracking players across an entire match using a single-camera setup. Our solution integrates a fine-tuned YOLOv5m for object detection, a SegFormer-based camera parameter estimator, and a DeepSORT-based tracking framework enhanced with re-identification, orientation prediction, and jersey number recognition. By ensuring both spatial accuracy and temporal consistency, our method delivers state-of-the-art game state reconstruction, securing first place in the SoccerNet Game State Reconstruction Challenge 2024 and significantly outperforming competing methods.
\end{abstract}

%% file: sec/1_intro.tex
\section{Introduction}
\label{sec:intro}

Game state reconstruction in football analytics represents a transformative step toward understanding and optimizing player performance, team strategies, and tactical decision-making during matches. By reconstructing the positions, roles, and identities of all individuals on the field—players, goalkeepers, referees, and others—coaches and analysts can derive actionable insights into player movements, team formations, and game dynamics. This capability is particularly critical in professional football, where precise tracking and role identification enable data-driven training strategies and enhance competitive advantage.

At its core, the SoccerNet Game State Reconstruction (GSR) challenge \cite{GSR} is a multiple object tracking (MOT) task, but with additional complexities that significantly elevate the difficulty. Beyond simply localizing athletes in 2D pitch coordinates, the task requires determining their roles (e.g., player, goalkeeper, referee), jersey numbers, and team affiliations (left or right relative to the camera viewpoint). These requirements introduce unique challenges, such as handling occlusions, distinguishing between players with similar appearances, and accurately associating identities across fragmented trajectories. Furthermore, the single-camera setup commonly used in football broadcasts introduces erratic camera movements, varying perspectives, and dynamic scene content, making accurate tracking and localization even more challenging.

Our work presents a robust end-to-end pipeline for tackling the SoccerNet GSR challenge, achieving state-of-the-art performance and securing first place in the 2024 competition. Key contributions of this paper include:
\begin{itemize}
\item A \textbf{Modular Pipeline Design} ensures high performance, flexibility, and seamless integration of new modules.
\item \textbf{Unique Camera Parameters Network}, which enables accurate mapping of detected objects from image-space to real-world pitch coordinates. 
\item \textbf{Pitch Localization with Keypoints Refinement}: Our pipeline enhances camera parameter predictions by detecting pitch line intersections and minimizing reprojection errors, ensuring precise and consistent athlete localization.
\item \textbf{Innovative Post-Processing}: We introduce a sophisticated post-processing stage that refines and merges short tracklets into longer, consistent trajectories, significantly reducing fragmentation and identity swaps.
\item \textbf{Superior Performance}: Our pipeline achieved a GS-HOTA score of \textbf{63.81}, significantly outperforming the second-place solution (43.15). 
\end{itemize}



%% file: sec/2_related_works.tex
\section{Related Works}
\label{sec:realted_work}

Since game state reconstruction was first introduced in 2024, few papers explicitly address the task. However, it can be decomposed into four critical subtasks: \textbf{player detection and tracking , pitch localization , team recognition , and jersey number detection}. 

\textbf{Players detection and tracking }in sports analytics is dominated by the tracking-by-detection paradigm, which typically combines one-stage object detectors like YOLOv8~\cite{yolov8} or other~\cite{centernet, gsr_4_bergmann2019tracking, gsr_6_bewley2016simple, gsr_7_bochinski2018extending, gsr_79_shitrit2023soccernet}\cite{gsr_81_somers2023body, gsr_86_thomas2017compute, gsr_94_ye2005jersey, gsr_95_yu2018comprehensive} with DeepSORT-like algorithms~\cite{deepsort, bytetrack}, augmented by pre-trained re-identification (ReID) feature extractors.
While these methods enable real-time performance due to their computational efficiency, they struggle with tracklet fragmentation in dynamic scenes and identity swaps in crowded scenarios. These issues stem from Kalman filter limitations in predicting non-linear motion  \cite{kalman_review} and the challenges of ReID in football, where uniform similarity, occlusions, and drastic pose variations degrade feature matching \cite{gsr_17_cioppa2020context, gsr_20_cioppa2022scaling, gsr_37_held2024x, gsr_68_pappalardo2019public, gsr_75_santos2023hierarchical, gsr_84_theiner2023tvcalib, gsr_89_vats2021multi}. An alternative approach, tracking-by-attention \cite{bytetrack, zeng2022motr}, integrates detection and tracking into a unified transformer-based framework, achieving state-of-the-art results on benchmarks like DanceTrack \cite{sun2022dancetrack} and MOTChallenge \cite{dendorfer2021motchallenge}. However, these methods prioritize spatial accuracy over temporal consistency, often neglecting smooth tracklet association, which is critical for sports analytics. 
Additionally, the computational complexity of transformer-based tracking makes it impractical for real-time applications.

\textbf{Pitch localization} is crucial for game state reconstruction, enabling the conversion of player positions from image space to real-world coordinates by estimating homography parameters.
The most common method relies on detecting correspondences between predicted keypoints \cite{chu2022sports_field_kpts, nie2021robust_field_kpts, jacquelin2022efficient_kpts_seg} or extracting field lines and circles using semantic segmentation \cite{homayounfar2017sports_field_seg, zhang2021high_cam_calib}, aligning them with a predefined field model using RANSAC \cite{fischler1981random_ransac}. While keypoint-based methods are computationally efficient, they suffer from instability in broadcast footage, where only a small subset of field markings is visible at any moment.

Alternative approaches estimate homography directly via CNN-based regression \cite{tarashima2020sflnet} or use search-based optimization, matching image features with a database of precomputed homographies \cite{chen2019sports_synt_cam_calib, sharma2018automated_top_view_field}. While these methods offer greater stability than keypoint-based approaches, they require large-scale databases for robust feature matching and still tend to be less precise and often require post-prediction refinement.

The refinement stage improves accuracy by minimizing reprojection error \cite{shi2022self_fiel_shape_alignment, gsr_84_theiner2023tvcalib, nie2021robust_field_kpts}, with temporal consistency enforced through smoothing techniques like Savitzky-Golay filtering \cite{gallagher2020savitzky_savgol} and outlier removal.

In \textbf{Team Recognition} task, most modern player-to-team assignment methods follow a pipeline similar to that described in \cite{GSR}, which consists of feature extraction and clustering. In the feature extraction stage, player appearance information is collected, often in the form of color histograms or Re-ID embeddings trained via contrastive learning \cite{mansourian2023multi_task_reid}. Extracted features are then clustered using unsupervised methods such as DBSCAN~\cite{ester1996density} or k-means~\cite{lloyd1982least}, grouping players into up to 2–5 clusters corresponding to referees, field players, and goalkeepers \cite{GSR}.
An alternative approach, \textbf{Associative Embedding (AE)}, eliminates the need for clustering by training a CNN to generate team-discriminative embeddings directly \cite{istasse2019associative_AE}. 
Unlike clustering-based pipelines, AE does not require explicit position-based assignment, making it more adaptable to different sports and camera views. 

\textbf{Jersey Number Detection} methods are broadly categorized into two approaches: (1) adaptations of general-purpose OCR techniques \cite{GSR, li2019show_ocr_baseline, liao2020_text_diff_binarization} and (2) custom solutions tailored for sports scenarios \cite{balaji2023jersey, koshkina2024_framework_jersey_numbers, gerke2015soccer_jersey, li2018jersey_semi_supervised, liu2019_jerseys_rcnn}. The first approach typically follows a two-step pipeline: text detection on localized player regions using methods like \textit{Differentiable Binarization} \cite{liao2020_text_diff_binarization}, followed by recognition via models such as \textit{Show, Attend and Read} \cite{li2019show_ocr_baseline}. While these methods offer development simplicity, their accuracy suffers in scenarios with distant players, occlusions, or sideways orientations. Custom solutions address domain-specific challenges by integrating sports-aware designs, such as one-stage number classification \cite{gerke2015soccer_jersey, li2018jersey_semi_supervised}, pose-guided detectors with post-processing \cite{liu2019_jerseys_rcnn}, or keyframe identification modules leveraging spatiotemporal networks to extract critical frames with visible numbers \cite{balaji2023jersey}. These methods boost accuracy using contextual features but need dataset-specific training, adding implementation complexity. 

%% file: sec/3_methodology.tex
\section{Methodology}
\label{sec:methodology}

\begin{figure*}
    \centering
    \includegraphics[width=1\linewidth]{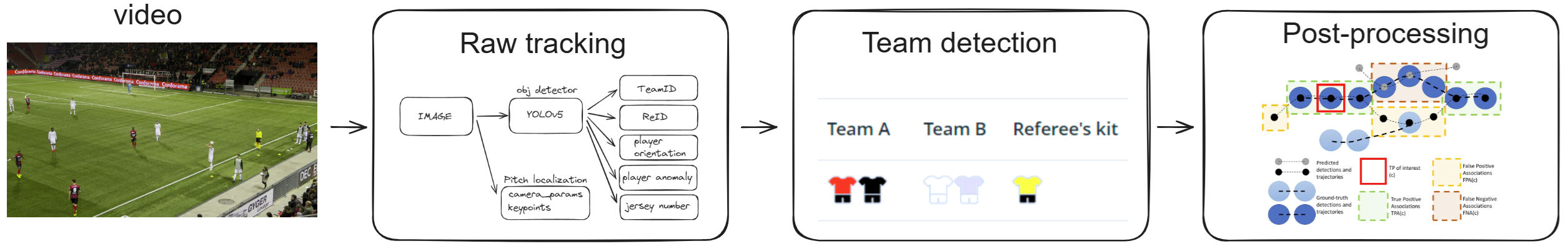}
    \caption{The overall pipeline is divided into three main stages. In the \textbf{raw tracking stage}, initial tracks are generated, and information about team embeddings and jersey numbers is estimated for each player. During the \textbf{team detection stage}, the previously collected information is used to assign player tracks to their respective teams. Finally, \textbf{postprocessing} is applied to reduce the number of resulting tracks by merging raw tracking results.}
    \label{fig:pipe_overview}
    \vspace{-1em} 
\end{figure*}

Our game state reconstruction framework is a robust and modular pipeline designed to track players and reconstruct their positions on a 2D top-view (minimap) of the football pitch using single-camera video footage (Figure \ref{fig:pipe_overview}). The tracking pipeline consists of three key stages: \textbf{raw tracking}, \textbf{team detection}, and \textbf{post-processing}.

In the \textbf{raw tracking} stage, athletes are detected, pitch localization is performed to determine their coordinates on the field, initial tracking is established, and feature vectors (ReID, TeamID, and jersey number) are computed for further processing. Next, in the \textbf{team detection} stage, all athletes on the field are grouped into five clusters based on their roles. Finally, in the \textbf{post-processing} stage, the system refines and merges short tracklets into longer, continuous trajectories by leveraging all previously gathered information.

\subsection{Raw Tracking}

\begin{figure}
    \centering
    \includegraphics[width=1\linewidth]{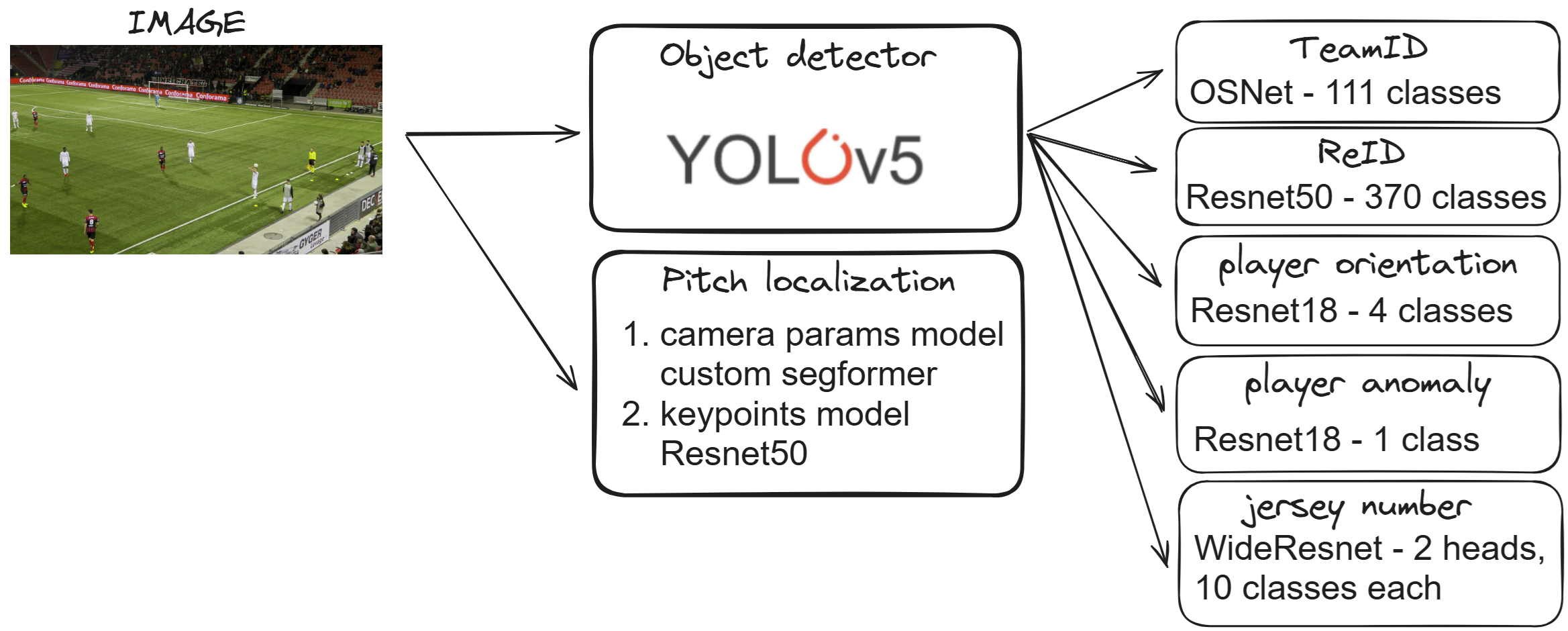}
    \caption{Raw tracking stage performs object detection, pitch localization, collects information about players teams required on consequent stages, Re-ID embeddings, jersey numbers and then merges all collected data into preliminary object tracks using the DeepSort-based tracking}
    \label{fig:raw_tracking}
    \vspace{-1em} 
\end{figure} 

Raw tracking is the first stage of our pipeline, generating preliminary tracking results in real time as video frames are processed (Figure \ref{fig:raw_tracking}). Since this stage is the most computationally intensive, we implemented it using NVIDIA DeepStream SDK \cite{nvidia2025deepstream}, optimizing all models as TensorRT engines \cite{nvidia2025tensorrt} with FP16 precision. This approach allowed us to achieve real-time tracking at up to 80 FPS on consumer-grade hardware, such as an RTX 3080Ti laptop GPU.

Firstly, for each input frame, the system detects players by YOLO detector \cite{yolov5}, fine-tuned for football-specific object detection, localizes athletes and the ball in the video frames. YOLOv5 architecture was chosen due to its one-stage detection strategy, relatively lightweight architecture which allows acceptable trade-off between quality and inference speed.

Then the proposed system proceed with \textbf{Pitch Localization}, which is a critical component of our pipeline, enabling the accurate mapping of detected objects from image-space to real-world pitch coordinates. This process is divided into two key stages: Initial Camera Parameter Prediction and Keypoints-Based Refinement (see Supplementary Material for a visualization of the entire process).

At \textbf{Initial Camera Parameters Prediction} a standard pinhole camera model was utilized, assuming zero distortions, and fixing the principal point to the center of the frame. The initial camera parameters for each frame are predicted using a custom CNN-Transformer encoder-decoder model. The encoder is based on the SegFormer architecture \cite{segformer}, while the decoder comprises multiple heads trained simultaneously. The model takes an H×W image from a football broadcast as input and predicts camera parameters.

The first head, the \textbf{“parameters head”} is trained to estimate seven camera parameters: position (x, y, z real-world coordinates), orientation (pan, roll, tilt), and field of view (FOV). This head incorporates a \textbf{Polarized Self-Attention (PSA)} layer \cite{liu2021_pca_layer} followed by multiple Conv2D layers and an average pooling layer.
The second head, the \textbf{“heatmaps head”} is trained to predict heatmaps for X and Y coordinates \cite{poranne2017_uv_mapping}. It is used only during training to enhance the training process by adding more modalities and distilling task-specific knowledge. This head consists of a PSA layer followed by Conv2D and Pixel Shuffle layers \cite{shi2016_sub_pix_shuffle} (further details about the model architecture can be found in Supplementary Material).

\setlength{\abovedisplayskip}{5pt} 
\setlength{\abovedisplayshortskip}{3pt} 

The proposed camera parameters model is trained with the following \textbf{loss function}:
\begin{multline}
\mathcal{L} = w_1 \cdot L^2_{\text{world}} + w_2 \cdot L^2_{\text{camera}}\\ 
+ w_3 \cdot L^1_{\text{parameters}} + w_4 \cdot L^2_{\text{heatmap}} 
\label{eq1}\tag{1}
\end{multline}
where \(w_1, w_2, w_3, w_4\) are the loss weights. \(L^2_{\text{world}}\) is the reprojection error in world space: 

\begin{equation*}
L^2_{\text{world}} = \sum_{x} \left\| P_{\text{pred}}^{\text{inv}}(P_{\text{gt}}(x_{\text{world}})) - x_{\text{world}} \right\|_2 
\label{eq2}\tag{2}
\end{equation*}
where \(P(x_{\text{world}})\) is a function that maps a 3D point in world coordinates \(x_{\text{world}}\) to 3D camera coordinates expressed in Normalized Device Coordinates (NDC), and \(P^{\text{inv}}(x_{\text{camera}})\) is its inverse function that maps NDC camera coordinates \(x_{\text{camera}}\) back to 3D world coordinates \(x_{\text{world}}\). The subscripts `\(\text{gt}\)` and `\(\text{pred}\)` indicate whether the function uses ground-truth or predicted camera parameters, respectively. 

\(L^2_{\text{camera}}\) denotes the loss in NDC camera space, allowing resolution-invariant learning and smooth gradient propagation due to the absence of the perspective divide.

\begin{equation*}
L^2_{\text{camera}} = \sum_{x} \left\| P_{\text{pred}}(x_{\text{world}}) - P_{\text{gt}}(x_{\text{world}}) \right\|_2
\label{eq3}\tag{3}
\end{equation*}

\(L^1_{\text{parameters}}\) represents the difference between predicted and real camera parameters (camera position, pitch, yaw, roll, field of view):  

\begin{equation*}
L^1_{\text{parameters}} = \sum_{y}\left\|(y_{\text{pred}} - y_{\text{gt}})\right\|_1
\label{eq4}\tag{4}
\end{equation*}
where \(y\) is a specific camera parameter. Term of the loss function~\cref{eq1}

\begin{equation*}
L^2_{\text{heatmap}} = \sum_{i,j} \left\| UV_{\text{pred}}(i,j) - UV_{\text{gt}}(i,j) \right\|_2
\label{eq5}\tag{5}
\end{equation*}
represents the difference between the ground truth heatmap \(UV_{\text{gt}}\) and the heatmap predicted by the network \(UV_{\text{pred}}\), where \(i, j\) are pixel coordinates on the heatmap (see details in Supplementary Material).

At \textbf{Keypoints-Based Refinement} stage the Field Keypoints Model, based on ResNet18 \cite{he2016_resnet}, detects 74 keypoints representing the intersections of pitch lines with grass lines. These keypoints refine the initial camera parameter predictions by minimizing the reprojection error when mapped to a canonical (two-dimensional aerial) field view.

This refinement process employs a brute-force optimization strategy, evaluating predefined combinations of camera parameter corrections and selecting the best one. A predefined set of delta values, such as $[-0.15, -0.10, \ldots, 0.15]$, is used for each camera parameter. These deltas are added to the SegFormer \cite{segformer} model's outputs, generating a delta matrix. Using these adjusted camera parameters, a series of homography matrices are computed to project the detected keypoints from the Field Keypoints Model onto planar coordinates (in the world space).

To ensure alignment, keypoints are converted into lines, where applicable, based on the expectation that certain keypoints lie along the same line. Outliers --- keypoints deviating significantly from the expected line are discarded. Multiple sets of lines are then generated, and the L2 distance between these planar lines and the ideal pitch model is calculated. The optimal camera parameters are determined by selecting the set with the smallest L2 difference, ensuring the most accurate alignment and projection.

The final camera parameters derived from this process enable the calculation of real-world 3D coordinates for all detected athletes. These coordinates are used by the DeepSORT algorithm \cite{deepsort} for tracking player movements on the pitch, providing a bird's-eye view of the game.

To reduce frame-to-frame fluctuations and ensure temporal smoothness in camera parameter predictions, a Savitzky-Golay filter~\cite{gallagher2020savitzky_savgol} is applied. This method efficiently smooths predictions but is limited to correcting small errors, with a maximum adjustment range of $\pm 2$ degrees for angles and $\pm 2$ meters for positions. To enable its use, predictions are delayed by 15 frames (0.5 sec), allowing the filter to access both past and future values.

The \textbf{Jersey Number Recognition} model is based on a modified ResNet architecture designed for 32×32 input images. Instead of detecting individual digits using bounding boxes, the model employs two classification heads: the first determines whether a jersey number contains a leading digit (1–9 or none), while the second predicts the second digit (0–9). This structure eliminates the need for precise bounding box annotations, simplifying data labeling and improving robustness.

To handle class imbalance in jersey number occurrences, a hybrid loss function was used in training: BinaryFocalLoss~\cite{lin2017focal} helps in determining the presence of a leading digit, while CrossEntropyLoss ensures accurate digit classification. This combination enhances performance on less frequent digit classes and improves overall robustness. The model processes each jersey crop independently and predicts two digits per image, allowing for recognition even when part of the number is occluded.


At \textbf{ Feature Extraction} stage secondary networks infer additional attributes for each athlete: \textbf{ReID Embeddings} are extracted using a ResNet50-based~\cite{he2016_resnet} model trained on athlete crops, enabling DeepSORT~\cite{deepsort} to maintain consistent player identities across frames. \textbf{TeamID Embeddings} are generated by an OSNet~\cite{zhou2019_osnet} model trained on 111 uniform classes, providing uniform-specific features for team detection in subsequent stages. Each class represents a unique uniform of the football players (for example red t-shirt and white shorts). OSNet\cite{zhou2019_osnet} is designed for person re-identification (\textbf{ReID}), utilizing multi-scale residual blocks and a unified aggregation gate (\textbf{AG}) to dynamically fuse spatial features, ensuring both efficiency and strong discriminative power. \textbf{Player Orientation} is predicted using a ResNet18~\cite{he2016_resnet} model trained to classify athletes’ directions (left/up/right/down). \textbf{Player Anomaly} is predicted by a lightweight custom model identifies and discards crops containing multiple athletes or irrelevant content to improve tracking quality.

DeepSORT-based algorithm \cite{deepsort} is used for \textbf{preliminary tracking} and operates in real-world pitch coordinates derived from the camera parameters instead of pixel coordinates. To ensure accurate and consistent tracking, restrictions based on player orientation (e.g., preventing 180-degree direction changes in one frame) and TeamID Embeddings are applied.

Raw Tracking stage lays the foundation for subsequent steps, enabling efficient, real-time tracking while providing essential features and initial tracking results for further refinement.

\begin{figure}
    \centering
    \includegraphics[width=1\linewidth]{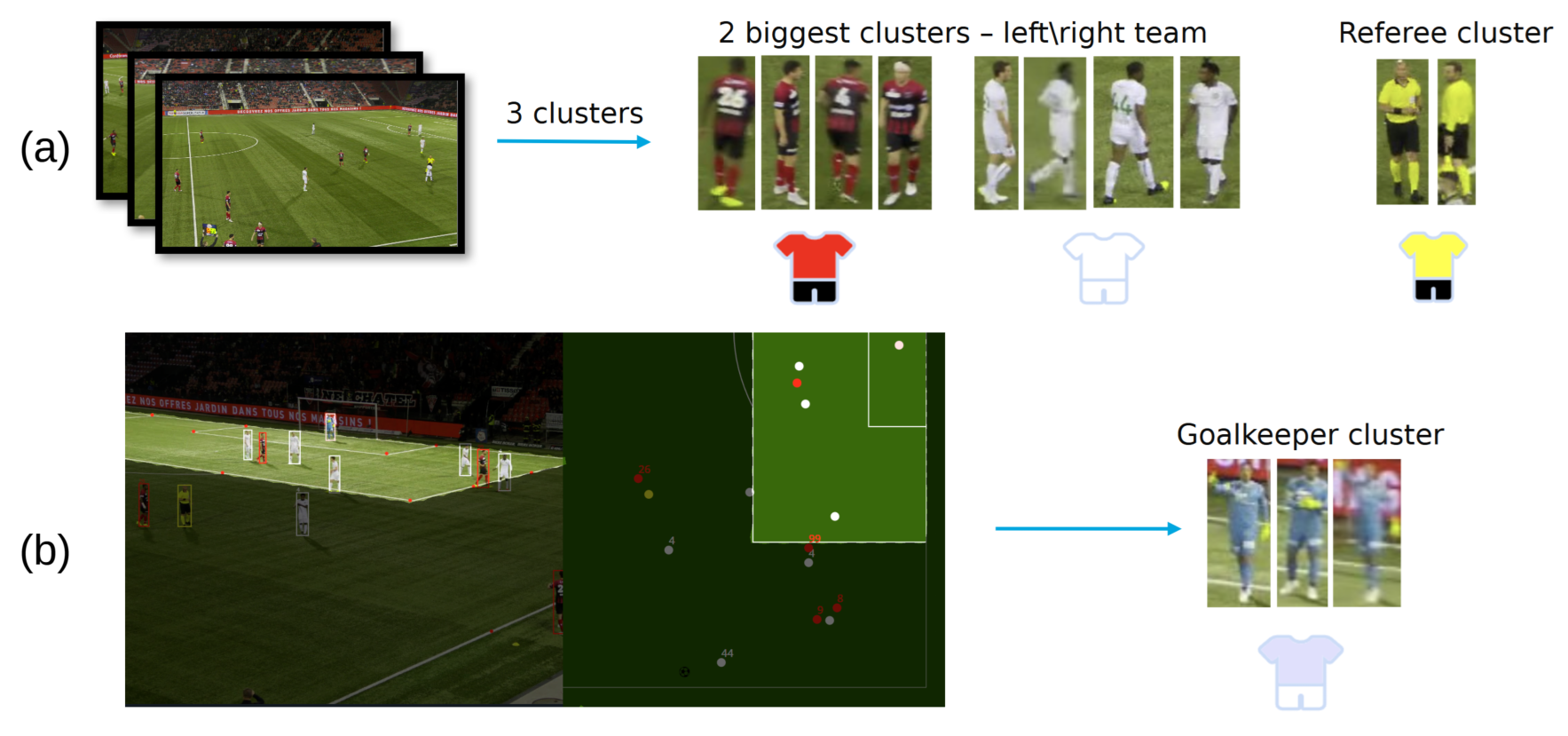}
    \caption{Team Detection Process. (a) Frames are clustered into three main groups: the two largest clusters (left and right teams) and the referee cluster. (b) Goalkeeper detection is performed separately by identifying athletes inside the penalty area and clustering them based on embeddings.}
    \label{fig:team_detection}
        \vspace{-1em} 
\end{figure}

\subsection{Team Detection}

In the previous step (raw-tracking), we gathered TeamID embeddings and pitch coordinates for every athlete across all frames. The next step is team detection , where we identify five embedding clusters: left/right team athletes, referees, and left/right goalkeepers . During post-processing, each athlete is assigned to a team based on the minimum cosine distance to the estimated clusters.

Estimating the Two Biggest Clusters (Left/Right Teams) and Referee Cluster (Figure \ref{fig:team_detection}a):
\begin{enumerate}
    \item We query all athletes whose X-axis distance is less than 30 meters from the pitch center and cluster them into three groups: the two largest clusters representing the left and right teams , and a smaller cluster for the referees.
    \item If the input video contains only the left or right penalty area (determined by the camera pan), or the clip is very short we instead select all athletes outside the penalty area . This strategy is not used by default, as goalkeepers occasionally leave their designated areas.
\end{enumerate}

\noindent Estimating Embeddings Clusters for Left and Right Goalkeepers (Figure \ref{fig:team_detection}b):
\begin{enumerate}
    \item We query athletes positioned inside the penalty area (on the left or right side).
    \item Athletes near previously detected clusters are filtered out based on cosine distance to avoid overlap.
    \item The remaining athletes are used to calculate an embedding cluster for the respective goalkeeper.
\end{enumerate}

\noindent\textbf{Defining Left vs. Right Teams}.\hspace{1em} To distinguish the left team from the right team , we employ a voting mechanism across frames (or at a defined stride for computational efficiency):
\begin{enumerate}
    \item For each processed frame, we calculate the mean X-coordinate of athletes belonging to one of the two largest clusters and collect votes indicating whether the cluster is positioned more to the left or right.
    \item After collecting all votes, the team affiliation is determined based on the majority vote count.
\end{enumerate}
This voting mechanism is more robust than simply calculating the mean X-coordinate across all frames, which can fail in very short clips or scenarios where only part of the football field is visible.

\subsection{Post-Processing}

The post-processing stage plays a critical role in refining and consolidating the raw tracking results generated during the preliminary tracking and team detection stages. By leveraging information such as player positions on the pitch, detected jersey numbers, re-identification (ReID) embeddings, and team labels, this stage addresses common tracking challenges, including tracklet fragmentation, identity swaps, and temporal inconsistencies. The primary objective of post-processing is to concatenate short tracklets into longer, coherent trajectories while ensuring consistency across all attributes.

The post-processing pipeline consists of four key stages: splitting tracklets by jersey numbers and team IDs, merging tracklets by jersey numbers, merging tracklets by ReID similarity, tracklet interpolation.



In the first stage, tracklets are split based on the assumption that each track should correspond to a single recognized jersey number and team label. This ensures that no tracklet contains conflicting or inconsistent attributes. After this step, all tracklets are guaranteed to contain only detections associated with a unique jersey number and team ID. This process effectively eliminates ambiguities arising from incorrect associations during earlier stages.

\label{sssec:jerseys_nums_merging}
The second stage focuses on merging tracklets that share the same jersey number. This is performed under the following conditions:
\begin{itemize}
    \item Temporal Non-Overlap : The tracklets must not overlap in time.
    \item Physical Feasibility : The distance between the end position of one tracklet and the start position of another must be physically traversable by a player within the time gap between the two tracklets.
    \item Consistent Team IDs : The team IDs of the tracklets being merged must match.
\end{itemize}
This stage significantly reduces fragmentation caused by temporary occlusions or misdetections.

For tracklets that remain unmerged after the jersey number-based merging stage, often due to missing or undetected jersey numbers caused by occlusions or unfavorable player orientations, we employ ReID feature similarity as an alternative criterion. The merging process adheres to the same foundational constraints outlined in Section \ref{sssec:jerseys_nums_merging}, ensuring temporal non-overlap, physical feasibility of movement, and consistency in team IDs.
To determine whether two tracklets belong to the same player, we leverage multiple ReID vectors along each tracklet. For example, consider Tracklet 1 with $N$ ReID vectors and Tracklet 2 with $M$ ReID vectors. We apply two complementary methods for comparing their ReID features:
\begin{itemize}
    \item \textit{Mean Vector Comparison :}
    For each tracklet, we compute the mean ReID vector by averaging all embeddings along the tracklet. The cosine distance between the mean vectors of Tracklet 1 and Tracklet 2 is then calculated. If this distance falls below a predefined threshold, the tracklets are merged. This approach provides a computationally efficient way to assess overall similarity.
    \item \textit{Cross-Multiplication Matrix (Pairwise Comparison) :}
    To achieve more robust matching, we construct an $N \times M$ matrix where each entry represents the cosine similarity between a pair of ReID vectors—one from Tracklet 1 and one from Tracklet 2. By identifying the maximum value in this matrix (or equivalently, the minimum cosine distance), we ensure that even noisy or inconsistent embeddings do not hinder accurate matching. This method effectively captures the most similar pair of embeddings across the two tracklets.
\end{itemize}
By combining these two strategies, our pipeline achieves higher accuracy in merging tracklets, particularly in challenging scenarios where individual ReID embeddings may be unreliable. This dual-strategy approach leverages both global (mean vector) and local (pairwise comparison) information, ensuring precise identity association across fragmented tracklets.

Finally, gaps in player trajectories are addressed using a classic linear interpolation algorithm. This step ensures smooth and continuous trajectories by filling in missing positions for brief periods when players temporarily leave the camera's field of view or are occluded.

The effectiveness of the post-processing pipeline depends heavily on the quality of the input data, including video resolution, weather conditions, and scene complexity. However, in general, the pipeline achieves a 90\% reduction in the number of tracklets and significantly minimizes tracklet swaps in crowded scenes. These improvements are particularly beneficial in challenging scenarios, such as dynamic content where players frequently enter and exit the camera's field of view.

By systematically addressing fragmentation, identity swaps, and temporal inconsistencies, the post-processing stage ensures high-quality, temporally consistent player trajectories, which are essential for accurate game state reconstruction.

%% file: sec/4_datasets.tex
\section{Datasets}
\label{sec:datasets}

All datasets used for training our models are summarized in \cref{tab:datasets}. These datasets cover a wide range of tasks, including object detection, re-identification, team classification, and pitch localization. While most datasets are briefly outlined in the table, the camera parameters dataset requires a more detailed explanation, which will be provided in a section below.

\subsection{Camera parameters dataset}

Our dataset mainly consists of two parts: real-world images (sourced from the Field Keypoints Dataset) and synthetic images (generated using a modified version of the Google Research Football Simulator~\cite{kurach2020google}). The combination of these datasets ensures a balance between real-world variability and controlled synthetic diversity, enhancing the robustness of the training process. Visualized distributions of camera parameters and example images can be found in Supplementary Material.

\noindent\textbf{Real-World Dataset.}\hspace{1em} To construct our dataset, we modified the TVCalib framework~\cite{gsr_84_theiner2023tvcalib} to work with labeled keypoints instead of segmentation masks. The dataset consists of images annotated with keypoints corresponding to the pitch lines in the image space.
Originally TVCalib calculates the loss in the image space. In our implementation, the optimization process operates in the real-world space (canonical view). Specifically, labeled keypoints from the image space are projected onto the canonical view, and the loss is defined as the sum of L2 distances between the projected keypoints and the corresponding pitch lines in the real-world space.
This adjustment to the framework proved to be more robust, leading to significantly more accurate camera parameter estimations compared to the original implementation. Using this process, we collected a dataset of 22,000 images with corresponding keypoint annotations.

\noindent\textbf{Synthetic Dataset.}\hspace{1em} To generate a synthetic dataset, we customized the Google Research Football Simulator~\cite{kurach2020google} to introduce randomized textures and camera views. Specifically, for every other launch of the simulator, textures for the grass, pitch lines, and athlete uniforms were selected randomly from a predefined set. This ensured a diverse appearance across the generated frames.
During each simulation, the game was played at 10x real-world speed, and camera parameters (e.g., position, orientation, and zoom) varied randomly as the game progressed. This approach allowed us to capture a wide range of perspectives on the football pitch efficiently.
To properly sample new camera positions while ensuring the camera still views the field, we performed sampling in multiple stages: the x-coordinate was uniformly sampled (in meters) from $-60$m to $60$m, the y-coordinate from $40$m to $110$m, and the z-coordinate from $-40$m to $-10$m; tilt and pan angles were adjusted to ensure the camera's center ray hit the football field, while the roll angle was fixed at 0 for all images, resulting in a dataset of 40,000 images annotated with camera parameters including position (x, y, z), orientation (pan, roll, tilt), and field of view (FoV). 

\setlength{\tabcolsep}{4pt} 
\begin{table*}
    \vspace{-1em} 
    \centering
    \resizebox{\textwidth}{!}{ 
        \begin{tabular}{@{}p{2.8cm} p{2.7cm} c c p{5.8cm}@{}}
            \toprule
            Task & Type & \# Imgs & \# Classes/kpts & Comment \\
            \midrule
            Detect athletes and ball & Object detection & 66k & 2 (athletes, ball) & YOLO-style dataset with bounding boxes \\
            Field Keypoints & Keypoint detection & 36k & 74 & Intersections of pitch lines with grass lines \\
            Camera Parameters & Regression & 62k & 7 (x, y, z, pan, roll, tilt, fov) & Includes synthetic and real-world data \\
            TeamID & Classification & 550k & 111 & Each class represents a unique team \\
            ReID & Re-Identification & 280k & 370 & Contains crops of 370 unique players \\
            Player Orientation & Classification & 20k & 4 (left, up, right, down) & Athlete facing direction classification \\
            Anomaly & Classification & 16k & 2 (normal, anomaly) & Detects irrelevant athlete crops \\
            Jersey Number Recognition & Classification & 70k & 100 & Upper half of athlete bbox with visible jersey number \\
            \bottomrule
        \end{tabular}
    }
    \caption{Datasets overview.}
    \label{tab:datasets}
    \vspace{-1em} 
\end{table*}

%% file: sec/5_experiment_setting.tex
\section{Experiment setting}
\label{sec:experiment_setting}

The training configurations for all models used in our pipeline are summarized in \cref{tab:training_config}, providing a comprehensive overview of the hyperparameters, loss functions, input sizes, and optimization details for each task. All models were trained using the Constructor Research Platform\footnote{\url{https://constructor.tech/products/research-platform}} with NVIDIA A100 GPU (40 GB). For camera parameter estimation we provide more in-depth description of the training setup. These additional details highlight the specific strategies and methodologies employed to achieve optimal performance for this task.

\setlength{\tabcolsep}{3pt} 
\begin{table*}[t] 
    \centering
    \resizebox{\textwidth}{!}{ 
        \begin{tabular}{@{}p{2.8cm} p{2.1cm} p{1.6cm} p{2.0cm} c c c p{6.5cm}@{}} 
            \toprule
            \textbf{Task} & \textbf{Type} & \textbf{Backbone} & \textbf{Loss} & \textbf{Input Size} & \textbf{mPara} & \textbf{Optimizer} & \textbf{Comment} \\
            \midrule
            Detect athletes and ball & Object detection & YOLOv5 & YOLOv5 loss (CIoU + BCE) & $1920\times1080$ & 35.5M & Adam & 100 epochs, LambdaLR scheduler, initial learning rate: $0.01$, momentum: $0.95$, decay: $5 \times 10^{-4}$ \\
            Field Keypoints & \parbox{2cm}{Keypoints \\ detection} & ResNet-18 & AdaptiveWing Loss & $480 \times 270$ & 11M & AdamW & 150 epochs, OneCycleLR, learning rate: $0.01$, learning rate factor: $0.1$ \\
            Camera Parameters & Regression & Custom SegFormer & Custom Loss & $512 \times 288$ & 5M & AdamW & 400k batches with batch size 8, cosine annealing scheduler, learning rate: $5 \times 10^{-4}$ \\
            TeamID & Classification & OSNet & TripletLoss & $64 \times 32$ & 0.3M & Adam & 40 epochs, learning rate: $0.001$, momentum: $0.9$, weight decay: $5 \times 10^{-4}$, single step learning rate scheduler with $\gamma =0.1$ and step size 1 \\
            ReID & Re-Identification & ResNet-50 & TripletLoss & $256 \times 128$ & 25.6M & SGD & Learning rate: $1 \times 10^{-3}$, triplet sampler: each batch contains 28 classes and 7 samples from each class \\
            Player Orientation & Classification & ResNet-18 & CrossEntropy & $62 \times 32$ & 11M & SGD & Learning rate: $1 \times 10^{-4}$ \\
            Anomaly & Classification & ResNet-18 & CrossEntropy & $62 \times 32$ & 11M & Adam & Learning rate: $1 \times 10^{-3}$ \\
            Jersey Number Recognition & Classification & ResNet-18 & BinaryFocal + CrossEntropy & $32 \times 32$ & 17M & AdamW & Learning rate: $1 \times 10^{-4}$ with Reduce LR on Plateau \\
            \bottomrule
        \end{tabular}
    }
    \caption{Training configurations.}
    \label{tab:training_config}
    \vspace{-1em} 
\end{table*}

\subsection{Camera parameters regression}
\noindent\textbf{Training setup.}\hspace{1em} The camera parameter regression model was trained in two distinct phases to ensure robust generalization across both synthetic and real-world data:
\begin{enumerate}
    \item The model was trained for 200k batches using a combination of real-world and synthetic datasets. This phase leveraged the diversity of synthetic data while grounding the model in real-world variability.
    \item An additional 200k batches were trained exclusively on real-world images to fine-tune the model for enhanced accuracy in practical scenarios.
\end{enumerate}
Training was conducted with a batch size of 8, utilizing the AdamW~\cite{loshchilov2017decoupled} optimizer. The learning rate was set to a maximum of $5 \times 10^{-4}$ and was dynamically adjusted using a cosine annealing scheduler with a warmup period of 20k steps to stabilize early training. The backbone weights were initialized using Kaiming initialization~\cite{he2015delving} to facilitate stable convergence.

\noindent\textbf{Hyperparameter Optimization.}\hspace{1em} To further enhance model performance, we employed the Optuna hyperparameter optimization framework~\cite{akiba2019optuna}. Optuna allows efficient exploration of the hyperparameter space using a tree-structured Parzen estimator (TPE) sampler. We optimized the weights \(w_1, w_2, w_3, w_4\) in the loss function~\cref{eq1} and the number of grid steps used for keypoint generation. This process involved randomly sampling weight values and grid configurations, followed by short training runs to identify the most effective combination. The final selected values were: \(w_1 = 0.048\) (\({L}^2_{\text{world}}\)), \(w_2 = 2.49\) (\({L}^2_{\text{camera}}\)), \(w_3 = 1.0\) (\({L}^2_{\text{params}}\), was fixed), \(w_4 = 10.0\) ( \({L}^2_{\text{heatmap}}\), introduced later and not optimized,), and \texttt{grid\_steps} = 36.


\noindent\textbf{Augmentations.}\hspace{1em} To enhance robustness, we applied two types of augmentations: \textbf{pixel-space} and \textbf{geometrical}, implemented using the Kornia library~\cite{riba2020kornia} for GPU-accelerated transformations.
\begin{itemize}
    \item \textbf{Pixel-Space Augmentations}: Modified image appearance without altering camera parameters, including cutouts, horizontal flipping, Gaussian noise/blur, sharpness adjustments, random color variations (contrast, brightness, hue, saturation), CLAHE, and shadow/brightness/contrast effects (e.g., RandomPlasmaShadow, RandomPlasmaBrightness). Each was applied with a probability of 0.5.
    \item \textbf{Geometrical Augmentations}: Affected both images and camera parameters, such as random rotations (adjusting the "roll" parameter) and horizontal flips (inverting the "pan" parameter).
\end{itemize}
These augmentations improved generalization across diverse broadcast conditions and camera setups.

%% file: sec/6_Evaluation.tex
\section{Evaluation (SoccerNet GSR)}
\label{sec:Evaluation}

We evaluated our method in the SoccerNet Game State Reconstruction (GSR) Challenge 2024~\cite{GSR}, which focuses on tracking and identifying players from a single-camera setup to generate a minimap representation of the game. The dataset consists of 200 video sequences (30 seconds each) with 2.36 million annotated athlete bounding boxes and 9.37 million pitch keypoints for localization. Ground truth labels for a subset of the data were kept private for evaluation on \textit{Eval.ai}\footnote{\url{https://eval.ai}}.  

\subsection{GS-HOTA Metric}

Performance was evaluated using GS-HOTA, a metric from ~\cite{GSR} tailored for game state reconstruction. GS-HOTA extends HOTA by factoring in role, team, and jersey number, ensuring practical applicability. It penalizes errors like incorrect attributes or missing labels, enforcing strict matching for high-quality results in football analytics. The similarity function includes:
\begin{itemize}
    \item Localization Similarity (LocSim) – Measures \textit{Euclidean distance} in pitch coordinates, smoothed by a Gaussian kernel with a 5-meter tolerance.  
    \item Identification Similarity (IdSim) – Requires \textit{exact matches} for role, team, and jersey number; mismatches are treated as false positives, enforcing strict tracking accuracy.  
\end{itemize}

\subsection{Results}


\begin{table}[h]
  \vspace{-1em} 
  \centering
  \label{tab:leaderboard}
  \resizebox{\columnwidth}{!}{
    \begin{tabular}{@{}l|c|c|c@{}}
      \toprule
      \textbf{Participant} & \textbf{GS-HOTA (↑)} & \textbf{GS-DetA (↑)} & \textbf{GS-AssA (↑)} \\
      \midrule
      \textbf{Constructor Tech (Ours)} & \textbf{63.81} & \textbf{49.52} & \textbf{82.23} \\
      UPCxMobius & 43.15 & 30.46 & 61.16 \\
      JAM & 34.40 & 19.38 & 61.08 \\
      \textit{Baseline}~\cite{GSR} & 23.36 & 9.80 & 55.69 \\
      \bottomrule
    \end{tabular}
  } 
    \caption{Game state reconstruction leaderboard. The main metric (GS-HOTA) and best performances are in \textbf{bold}.}
    \vspace{-1em} 
\end{table}

Our method secured \textbf{first place} in the SoccerNet GSR Challenge 2024, achieving a GS-HOTA score of \textbf{63.81}, significantly outperforming the second-place team (43.15) and third-place team (34.40). The comparison of our results with the top results of SoccerNet GSR Challenge 2024 and the baseline is given in \cref{tab:leaderboard}. 

The post-processing stage played a crucial role in improving association accuracy (GS-AssA) by reducing tracklet fragmentation, while our pitch localization pipeline enhanced detection accuracy (GS-DetA) by ensuring precise player positioning on the pitch. These improvements led to our significant performance margin over competing methods.  

%% file: sec/7_Conclusion.tex
\section{Conclusion and Future Work}
\label{sec:Conclusion}
We evaluated our approach on the largest football broadcast Game State Reconstruction dataset available, securing the top position in the SoccerNet Game State Reconstruction Challenge 2024. This achievement highlights the effectiveness of our method in real-world scenarios.
For future improvements, we aim to unify the camera models and field keypoints model by adding an additional head to the custom SegFormer. Additionally, we plan to replace YOLOv5, remove the anomaly model due to its negligible impact, and enhance player orientation modeling by training on a uniform angle distribution (0 to 360 degrees) instead of four discrete bins.
Furthermore, since our model predicts jersey numbers as separate first and second digits, we can relax the tracklet merging criteria to require only one matching digit (depending on the athlete’s orientation). When combined with the improved player orientation model, this adjustment could significantly enhance the association ability of our tracking pipeline, leading to more robust and accurate game state reconstruction.

%% file: sec/8_Acknowledgments.tex
\section{Acknowledgments}
\label{sec:Acknowledgments}
The authors would like to thank Igor Rekun for his valuable contributions to the development of the Camera Parameter model.

%% file: sec/X_suppl.tex
\clearpage
\appendix
\setcounter{page}{1}
\maketitlesupplementary

\section{Camera Parameters Network Architecture}
\label{sec:cam_param_arch}

\begin{figure*}
    \centering
    \includegraphics[width=1\linewidth]{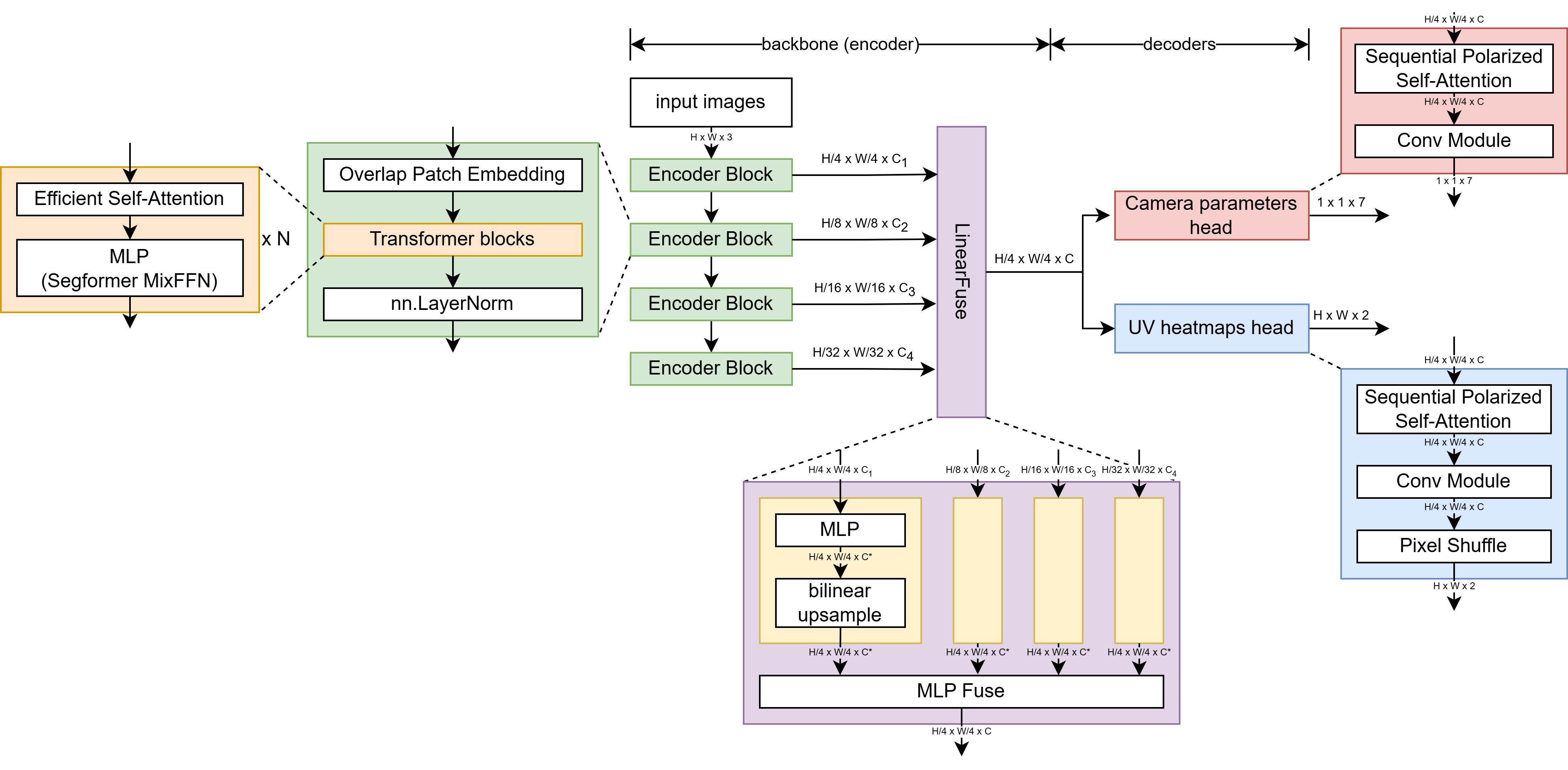}
    \caption{Camera Parameters Model. This figure illustrates the architecture of our custom SegFormer-based camera parameter estimator. The model consists of an encoder-decoder structure, where the encoder is based on the SegFormer architecture and the decoder includes two heads: one for predicting camera parameters (position, orientation, and field of view) and another for generating UV heatmaps.}
    \label{fig:cam_params_net}
\end{figure*}

See Figure~\ref{fig:cam_params_net} for a detailed diagram of the camera parameters prediction model architecture.

\section{Pitch Localization}

Accurate pitch localization is essential for mapping athletes on the image to their 3D positions on the pitch. Our method employs a multi-stage approach that first leverages a custom SegFormer model to generate an initial estimate of the camera parameters. Following this, a ResNet50-based segmentation network detects keypoints on the field, such as intersections of pitch lines with grass lines. These keypoints are then used in an optimization process to refine the estimated parameters, ensuring a more precise camera alignment with the real-world pitch.

\begin{figure*}
    \centering
    \includegraphics[width=1\linewidth]{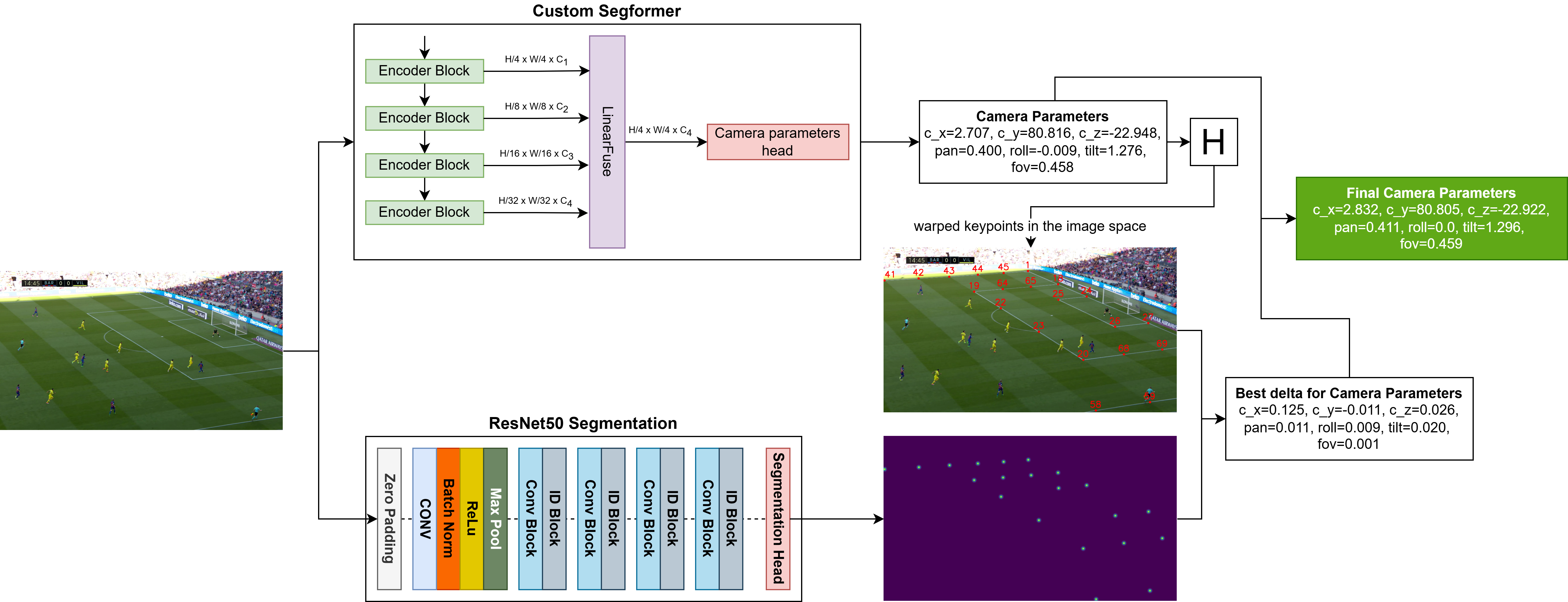}
    \caption{The pipeline estimates camera parameters by combining a custom SegFormer model for initial predictions and a ResNet50-based segmentation for keypoint detection. The parameters are refined using keypoint alignment to obtain the final camera pose.}
    \label{fig:field_det_pipe}
\end{figure*}

You can find visualization of the pitch localization pipeline in Figure~\ref{fig:field_det_pipe}.

\section{Camera Parameters Dataset}

\begin{figure*}
    \centering
    \includegraphics[width=1\linewidth]{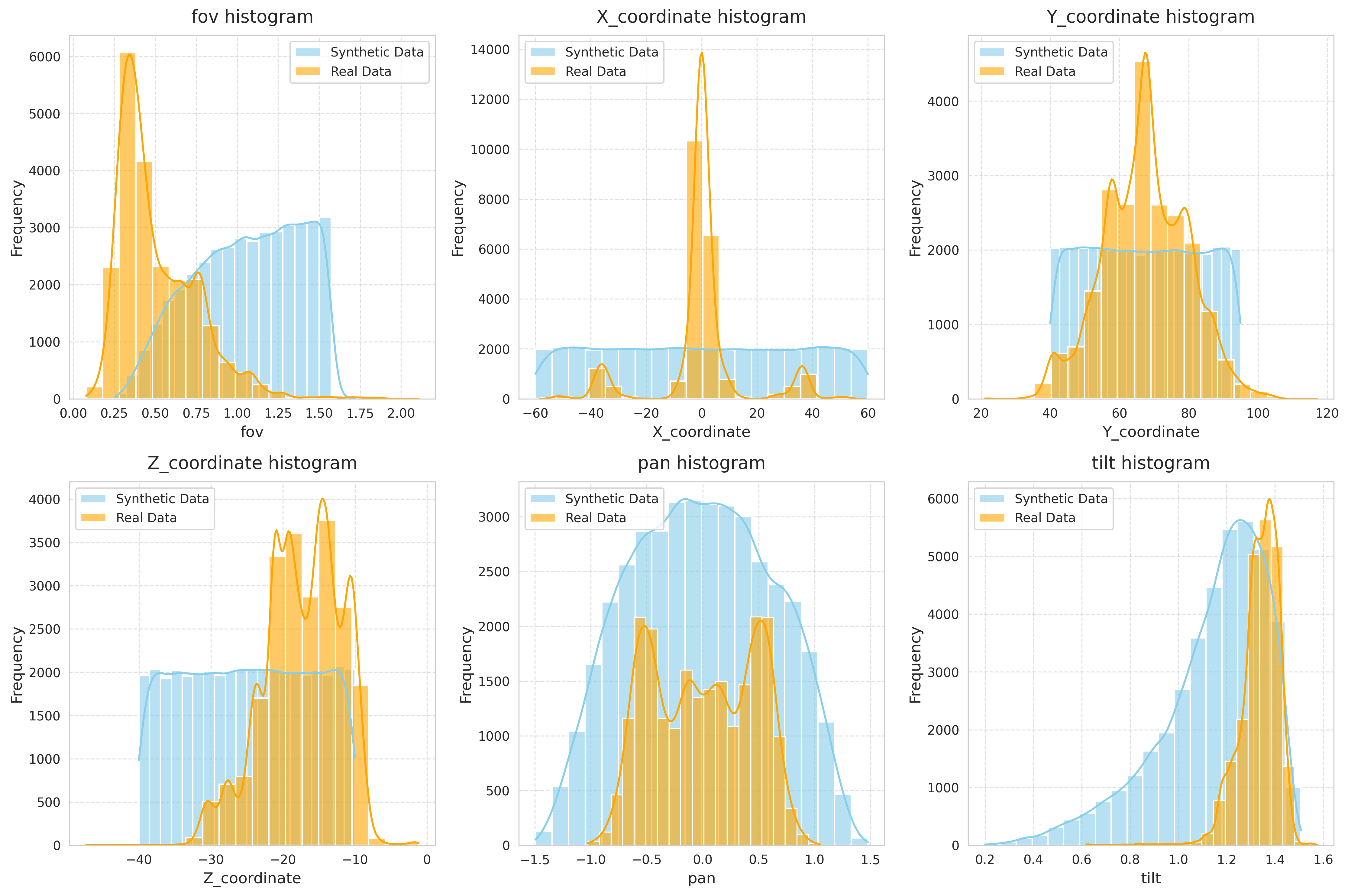}
    \caption{Real and synthetic data statistics. The histograms compare the distributions of key camera parameters and coordinate values between real and synthetic datasets. The X, Y, and Z coordinates represent camera positions with respect to the center of the football field, which serves as the origin (0,0,0). The FIFA standard field dimensions are 105 meters (length) × 68 meters (width). The field of view (FoV), pan, and tilt angles illustrate differences in camera configurations across datasets, while roll is fixed at 0 for all images. The synthetic data (blue) shows a broader and more uniform distribution, while the real data (orange) exhibits a more concentrated range of values, indicating the constrained nature of real-world camera placements.}
    \label{fig:dataset_statistics}
\end{figure*}

\begin{figure*}[h]
    \centering
    \begin{subfigure}[b]{0.48\linewidth}
        \centering
        \includegraphics[width=\linewidth]{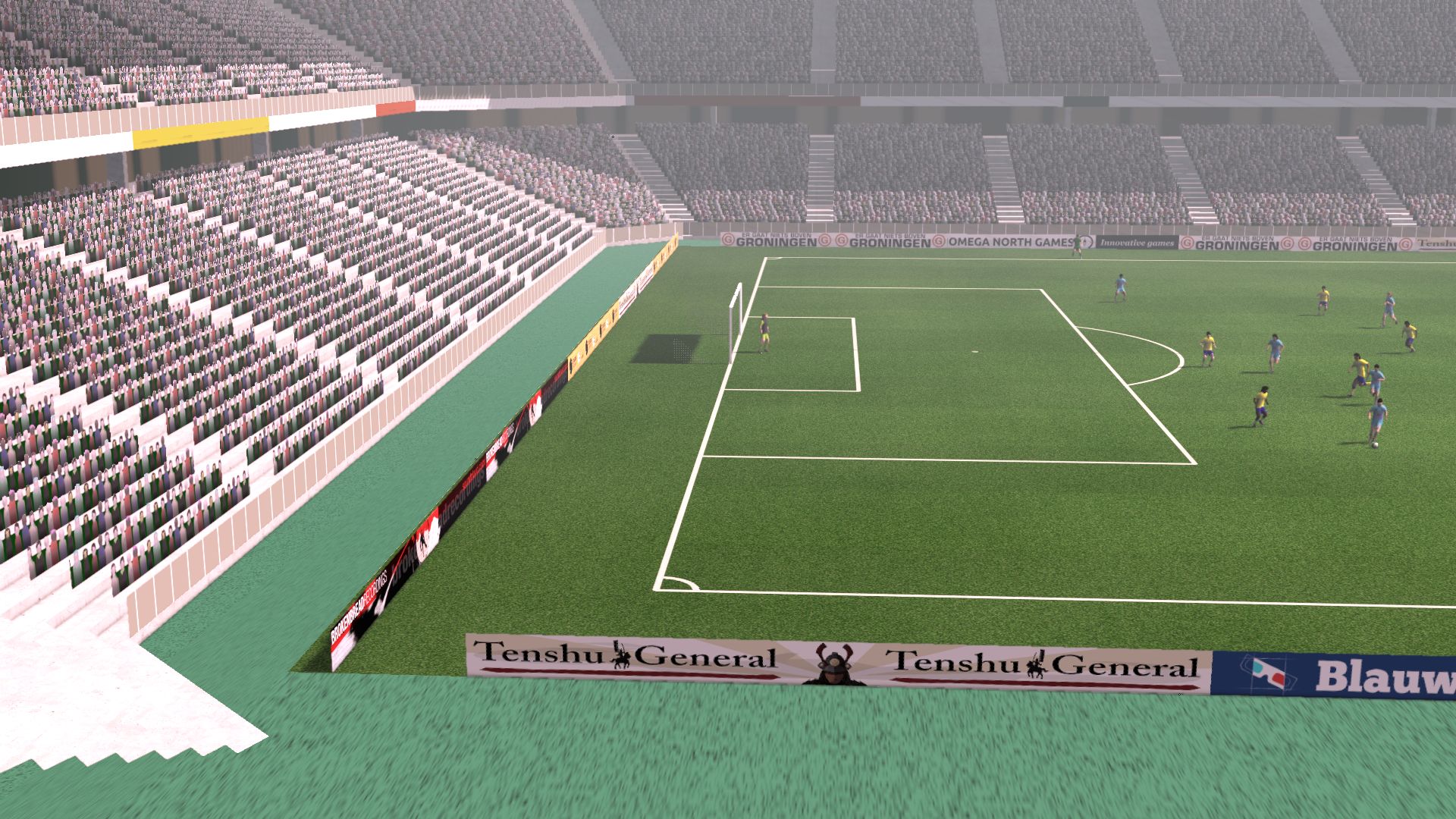}
        \caption{
            \small FoV: 0.86, 
            $c_x$: -48.19, $c_y$: 72.27, $c_z$: -13.31, 
            Pan: -0.06, Tilt: 1.35, Roll: 0.0
        }
    \end{subfigure}
    \hfill
    \begin{subfigure}[b]{0.48\linewidth}
        \centering
        \includegraphics[width=\linewidth]{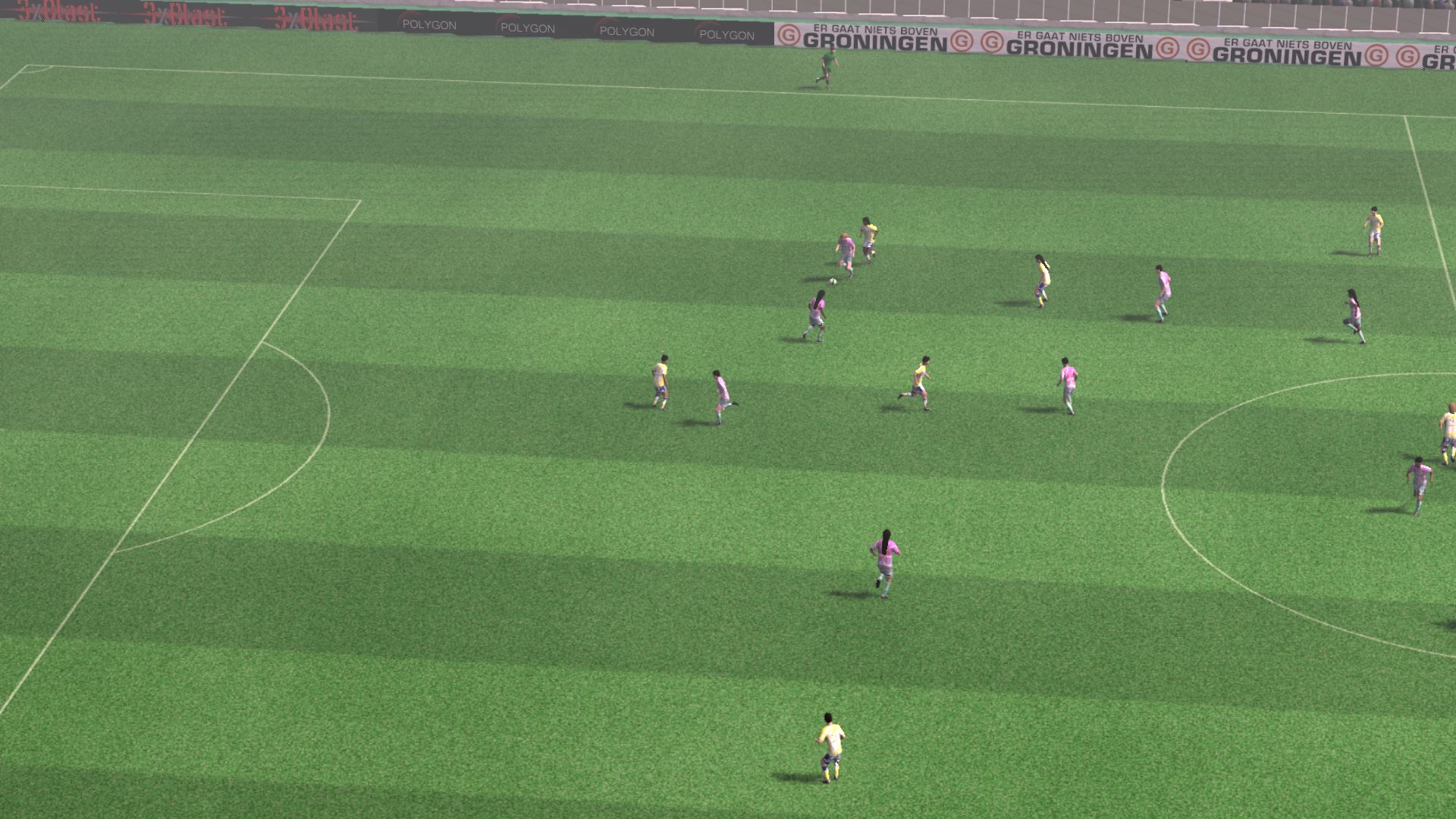}
        \caption{
            \small FoV: 0.47, 
            $c_x$: -11.75, $c_y$: 74.62, $c_z$: -34.18, 
            Pan: -0.12, Tilt: 1.16, Roll: 0.0
        }
    \end{subfigure}
    
    \vspace{0.5em}
    
    \begin{subfigure}[b]{0.48\linewidth}
        \centering
        \includegraphics[width=\linewidth]{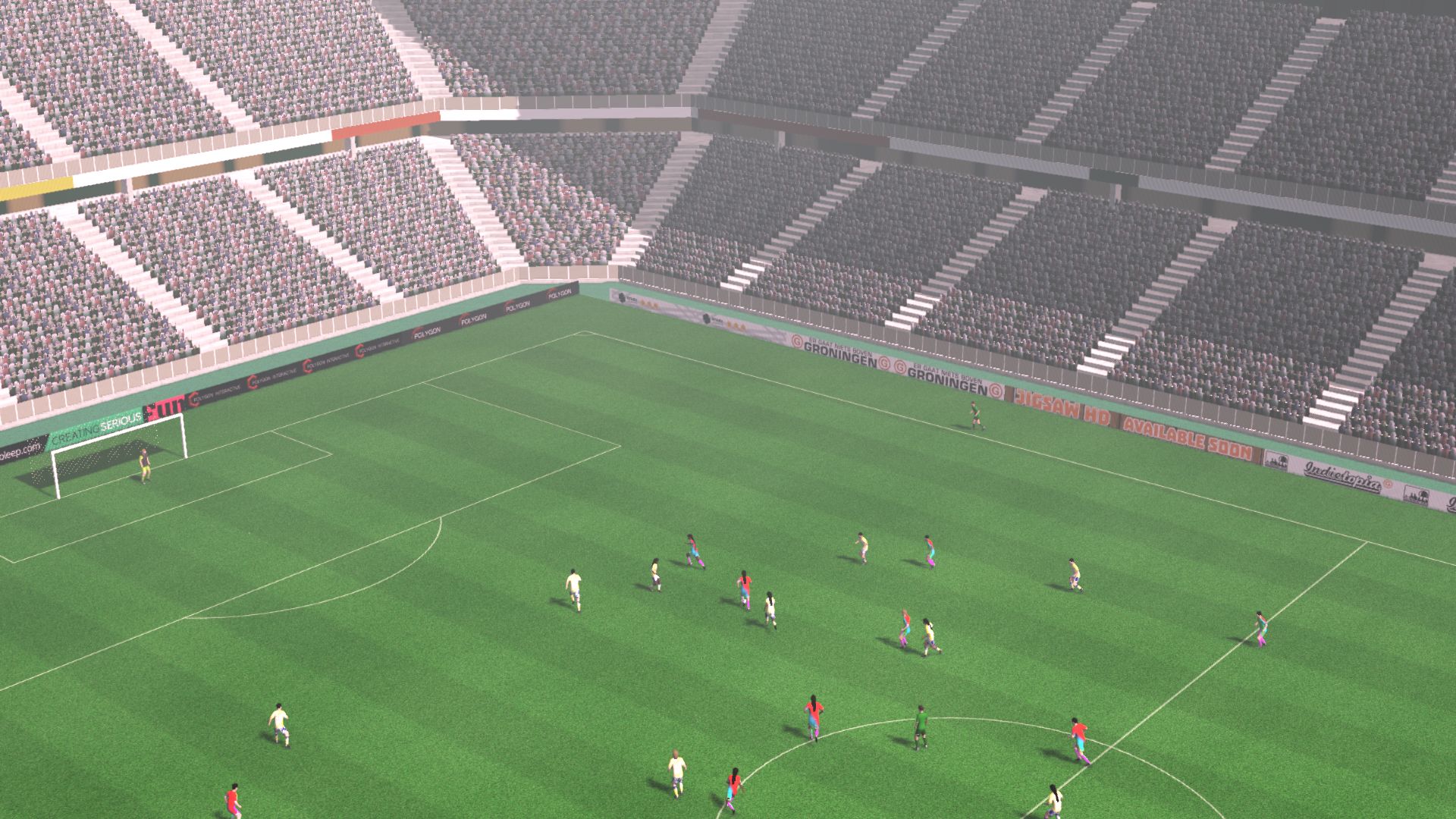}
        \caption{
            \small FoV: 0.78, 
            $c_x$: 26.77, $c_y$: 44.59, $c_z$: -34.42, 
            Pan: -0.71, Tilt: 1.23, Roll: 0.0
        }
    \end{subfigure}
    \hfill
    \begin{subfigure}[b]{0.48\linewidth}
        \centering
        \includegraphics[width=\linewidth]{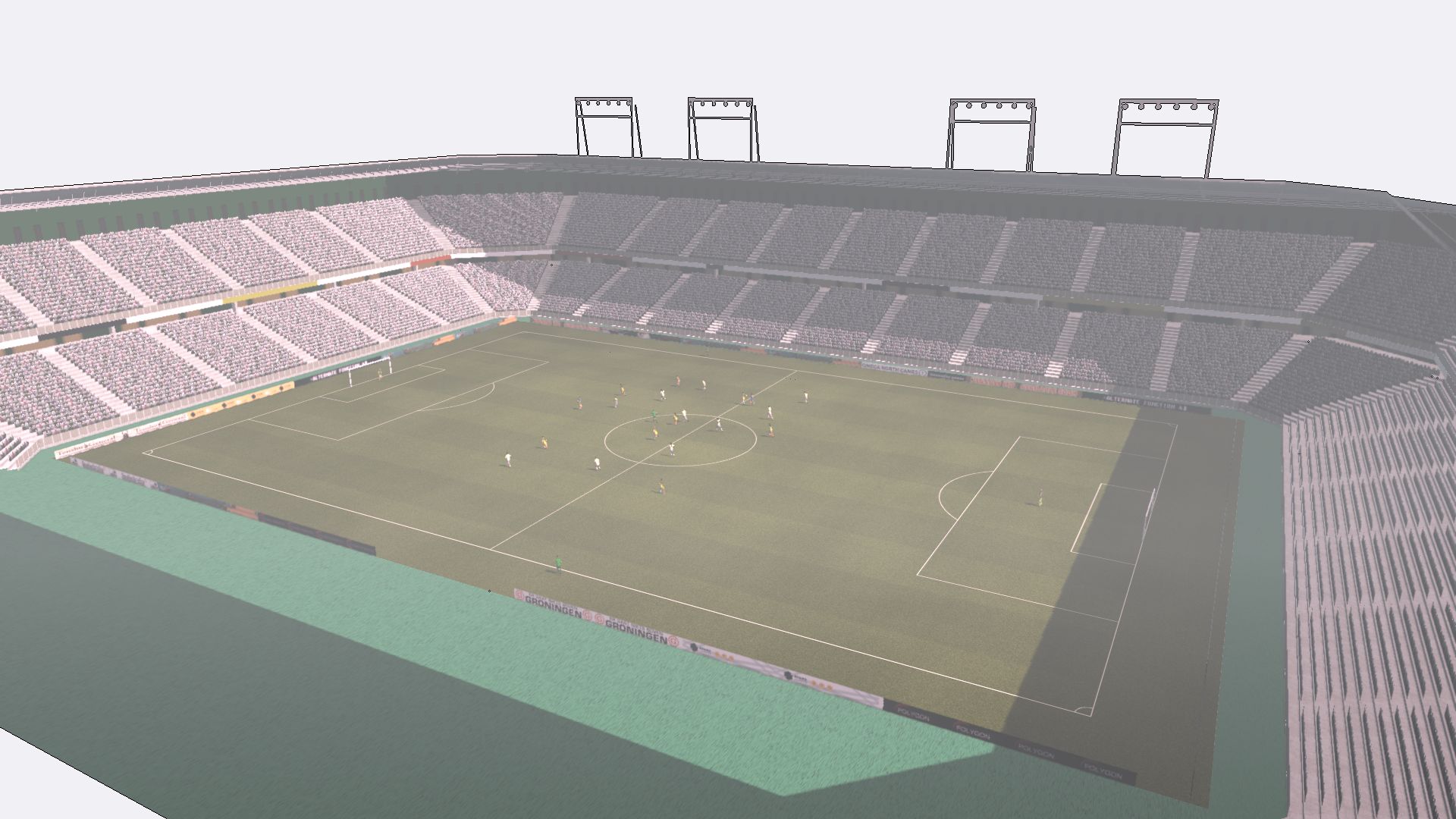}
        \caption{
            \small FoV: 1.29, 
            $c_x$: 57.28, $c_y$: 94.18, $c_z$: -39.87, 
            Pan: -0.49, Tilt: 1.25, Roll: 0.0
        }
    \end{subfigure}
    
    \caption{Examples from the synthetic dataset with corresponding camera parameters.}
    \label{fig:synthetic_examples}
\end{figure*}

The histograms in Figure~\ref{fig:dataset_statistics} compare key distributions, such as camera position coordinates (X, Y, Z), field of view, pan, and tilt angles. The real dataset is naturally constrained by physical camera placements, while the synthetic dataset spans a wider range of configurations, enabling the model to learn robust representations. 
You can see examples of synthetic images in Figure~\ref{fig:synthetic_examples}.

\section{Camera Parameters Loss}

Given the camera parameters \(\text{params} = \{I, R, t\}\), we define a mapping function \(P = F(\text{params})\) that transforms 3D world coordinates \(X\) into 3D camera coordinates in Normalized Device Coordinates (NDC) space:

\[
x_{\text{camera}} = P(X) = IR(X - t),
\]

where:
\begin{itemize}
    \item \(X\) is the 3D world coordinates \([X, Y, Z]^T\).
    \item \(I\) is the intrinsic matrix, encoding focal length and principal point (which is set to zero in the case of NDC coordinates).
    \item \(R\) is the rotation matrix representing the camera's orientation.
    \item \(t\) is the translation vector representing the camera's position.
    \item \(x_{\text{camera}}\) is the resulting 3D point \([x_c, y_c, z_c]^T\) in NDC space.
\end{itemize}

This 3D-to-3D transformation (from world coordinates to Normalized Device Coordinates, or NDC) offers three key advantages: (1) it ensures resolution invariance by decoupling the loss from the input image size, (2) it eliminates the need for a perspective divide, thereby maintaining a smooth and stable gradient flow during optimization, and (3) it retains invertibility, enabling consistent reconstruction of 3D world coordinates from camera coordinates.

The inverse mapping is derived as follows. Starting from the forward transformation:

\[
x_{\text{camera}} = IR(X - t),
\]

we derive the inverse as follows:

\begin{itemize}
    \item Multiply both sides by \((IR)^{-1}\) to isolate \(X - t\):
    \[
    (IR)^{-1} x_{\text{camera}} = X - t.
    \]

    \item Solve for \(X\) by adding \(t\) to both sides:
    \[
    X = (IR)^{-1} x_{\text{camera}} + t.
    \]

    \item Since \(R\) is an orthogonal matrix, \(R^{-1} = R^T\), giving the final expression:
    \[
    X = R^T I^{-1} x_{\text{camera}} + t.
    \]
\end{itemize}

Thus, our inverse mapping is:

\[
P^{\text{inv}}(x_{\text{camera}}) = R^T I^{-1} x_{\text{camera}} + t.
\]

This formulation plays a critical role in our training loss, allowing symmetric penalization of both forward and inverse transformations between world and NDC camera coordinates.

\begin{figure*}
    \centering
    \includegraphics[width=1\linewidth]{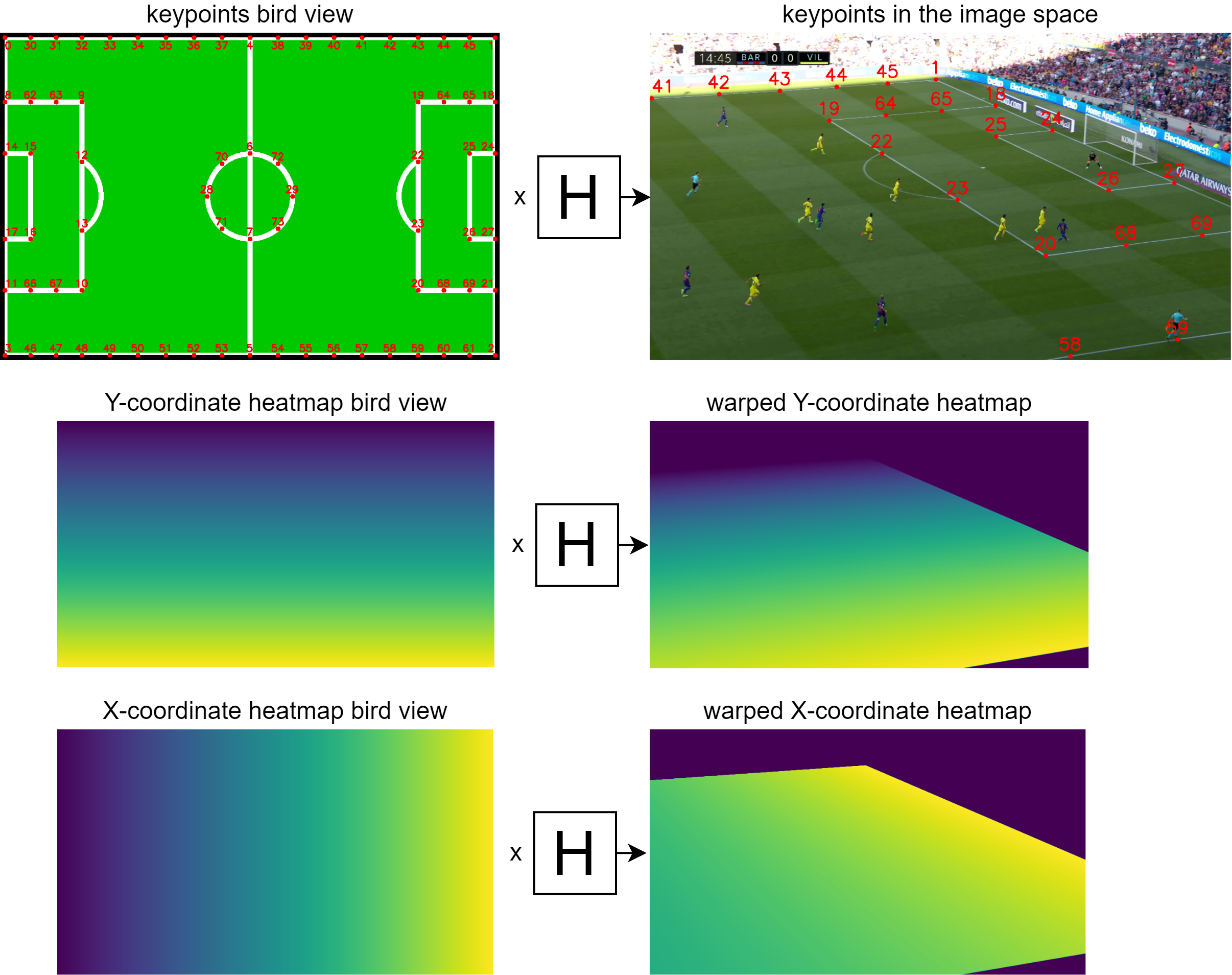}
    \caption{Keypoints, Y-coordinate heatmap (bird's-eye view), and X-coordinate heatmap are projected into image space using a homography matrix derived from the camera parameters.}
    \label{fig:UV_preparation}
\end{figure*}

\section{Camera parameters data preparation}

Figure~\ref{fig:UV_preparation} illustrates the projection of keypoints and coordinate heatmaps into image space using a homography matrix computed from the camera parameters.